# Studying the control of non-invasive prosthetic hands over large time spans

*Abstract*. The Electromyography (EMG) signal is the electrical manifestation of a neuromuscular activation that provides access to physiological processes which cause the muscle to generate force and produce movement. Non-invasive prostheses use such signals detected by electrodes placed on the user's stump, as input to generate hand posture movements according to the intentions of the prosthesis wearer. The aim of this pilot study is to explore the repeatability issue, i.e. the ability to classify 17 different hand postures, represented by EMG signal, across a time span of days by a control algorithm. Data collection experiments lasted four days and signals were collected from the forearm of a single subject. We find that Support Vector Machine (SVM) classification results are high enough to guarantee a correct classification of more than 10 postures in each moment of the considered time span.

# I. Introduction

The rapid evolution of technology and robotics can be seen more and more in everyday life. The great power of portable sensors and mechatronic technology stands out when these two are combined together in the prosthetics devices field.

The application of the latest research discoveries could greatly impact the quality of life of impaired people, but still a lot has to be done. According to previous studies (see [8] and references therein), controlling prosthesis is far away from natural motion and must undergo complicated and painful training sessions. Although some steps in this direction have been made [6], in most cases the tasks that a prosthesis can perform are limited to opening and closing, as the methods used to control such advanced hands are usually rudimentary, relying on sequential control strategies [9].

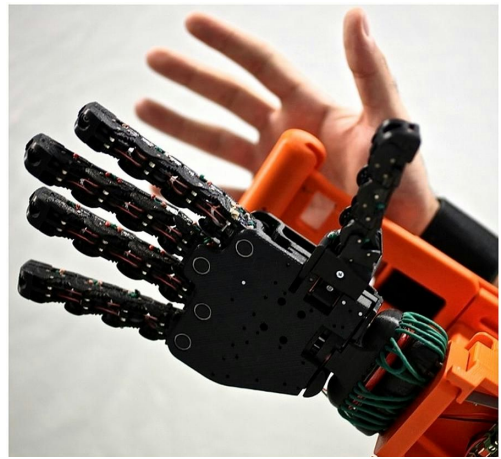

FIG.1 *The Pisa SoftHand and the forearm adapter used to test the device. Photo credits to [8].*

While excellent results have been obtained with invasive methods, non-invasive studies [1, 2] usually show average classification accuracies of hand movements up to 80–90%.

Thus, the movement recognition accuracy is never high enough to avoid misclassification on a large number of movements [1]. As a result, one of the main issues of prosthetic control is the training time needed by a user to alleviate the inconsistencies between the desired and performed movement. This process can take up to several days and is generally tiring and painful. In order to avoid a consequent withdrawal, this issue calls for machine learning techniques able to boost the learning process of each user.

Due to fatigue or electrode displacement and personal quantity of subcutaneous fat, skin and muscles qualities, each user needs a long training time before being able to fully exploit the prosthesis. The number of cumbersome training sessions could be dramatically reduced using an adaptive system that is already informed about the possible basic hand movements [8].

In this work we discuss the results of experiments in learning algorithms repeatability that attempted to highlight the variation in measurements and consequent deterioration of performance over long time spans. Shift in the electrodes caused by disconnection and attachment of the myoelectric device during the night and and other factors could be responsible for this deterioration.
A primary advantage of this study is that it demonstrates that offline learning methods themselves are not sufficient to a proper classification of movements. Moreover, it clearly shows the emergence of a trend in the performances obtained over long time spans. Once noticed this, a lot of work and corrections could be done.

The present report is organized as follows. Section II presents an overview of related works on this issue. Section III gives a problem definition, while materials and methods description can be found in Section IV, including the description of the data acquisition setup and the experimental protocol. Section V reports the experiments and the discussion of the results. Finally, Section VI draws conclusions and contains possible directions for future work.

# II. Related Work

One of the main issues for the seeking of dexterity in the control of advanced prosthesis is increasing the number of active degrees of freedom offered by the devices [8]. To make an example, with only one active degree the hand can open and close on command and nothing more [6]. However, in 2009 C.Castellini and P. van der Smagt showed that as many as nine different postures could be classified with a remarkable degree of accuracy [12]. Moreover, in recent studies such as the one for NinaPro Database the number of explored gestures and their consequent classification has been extended considerably. They have trained and tested the most common classifiers in machine learning with 52 movements and postures, including grasping and exerted force recognition, showing that classification of a large number of different hand tasks through machine learning algorithms is possible [1, 2, 9].

The use of surface EMG signals can be influenced by several factors not related to finger movements, that clearly may disturb final classification [6]. For instance, electrode conductivity changes, such as perspiration or humidity, or electrophysiological changes (muscle fatigue) or even spatial changes, due to movement on the skin or soft tissue fluid fluctuations, can vary, according to previous evaluations, the strength of the measured signal. Many more factors that might influence acquisitions can be found in [13]. As a consequence, inter-subject variability, electrode displacement and muscle fatigue should be taken into account when working in this field [12].

Being all forearms different in shape, size and power, experiments should at first be concentrated only on one subject and then extended to a higher number of them in order to achieve statistical significance. In addition, the intensity and quality of the EMG signal depend upon a correct placement of the electrode right over the muscle belly, but logically displacing the electrodes always at the very same position is impossible [13]. Not to be underestimated is that, as the muscles are used repeatedly during experiments, muscle fatigue becomes increasingly more perceivable in the subject thus influencing the measurements.

These issues has been vastly investigated in the literature. The only possibility to overcome these problems is to explicitly take them into account, gathering enough data to be able to train the machine under different conditions of electrode displacement and muscular fatigue [12].

Besides finger movements and grasping, the forearm muscles are also involved in the motion of the arm. The EMG signal is therefore likely to change if the forearm is moved during acquisitions, for example when switching from pronation to supination or simply while walking around. To overcome this inconvenient, in all previous studies the subject has been instructed to keep the arm still and relaxed on a table in comfortable position with the palm orthogonal to the plane. It has been suggested however, to take into into account even the changing arm posture by sampling more of the input space [12]. For this reason, a solution to this problem could be extending the 17 movements considered in this work to all the 52 movements exploited in literature.

Pattern classification of myoelectric signals has been widely investigated with promising results in the laboratory setting [5]. Selection of appropriate features has been driven by expert knowledge to a wide range of successful feature sets that have yielded low classification error on able bodied subjects and non.
The extraction methods that have been successful in this field consider spectral and amplitude properties of the signal and can be categorized in those operating in time domain or frequency domain. Mean Absolute Value (MAV), Variance (VAR), Waveform Length (WL) can be considered as an example for the first one. On the other hand, Frequency Ratio or Mean Frequency can be exhaustive as a representation of the second category (see [1] for features evaluation). Moreover, kind of sophisticated features in time-frequency domain can be considered. These ones are richer in details but highly more expensive in computation time. As an example, there are Short-Time Fourier Transform (STFT) and Wavelet Transform (WL).

According to the literature, the use of classification methods has mostly been restricted to relatively standard methods, such as Linear Discriminant Analysis, k-nearest neighbors or multi-layer perceptrons. Recently Support Vectors Machines have been extensively used in machine learning with many biomedical signal classification applications [12, 6]. In fact, under a discrete number of conditions, SVMs report a similar performance to Linear Discriminant Analysis [1].

Although these studies report low errors in a single session, robustness over time or across different activities has rarely been evaluated.
With this work we aim to do a pioneer investigation on repeatability of sEMG signals over large time spans taking into account materials and methods that previously were shown to be successful.

*FIG.2 Final ambitious goal representation. Photo credits to [8]*

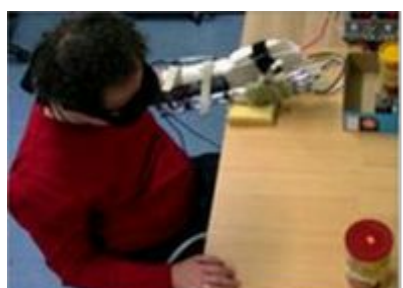
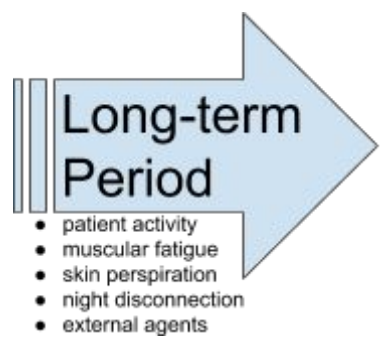
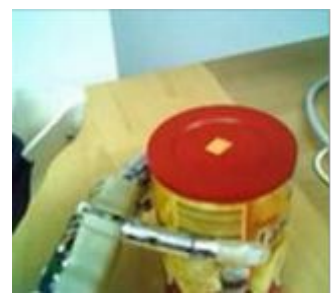

# III. Problem Statement

Repeatability is the measure of change in the variance of an operation performed repetitively under controlled conditions. Calculating repeatability can help researchers to found malfunctioning in prosthetic hands and their possible causes.

Suppose to have an advanced prosthetic hand that only needs to be fitted on the amputated limb each morning and that requires a single 30 minutes training session per week to correctly classify the desired movements. Using the prosthesis in real life conditions implies several changes to electrode displacements and their consequent measurements. To make an example, throughout the day sensors may shift a little. Reasonably, electrodes cannot be expected to exactly lie in the very same position each time the prosthesis is used. Even more relevant, it is not possible to guarantee an extremely precise reapplication of the prosthesis sensors in the same position each day, especially because the amputee should be able to apply it autonomously. Moreover, as the muscles are used continually, one can expect that fatigue might change the EMG signal. This should require a continual adaptation of the prosthesis to the subject condition.

The aim of this pilot study is to see how classification results change throughout the week, when training the prosthetic hands only on the first day. We want to see difference weight in accuracy after 1 day, 2 days .. seeing whether this difference increases in a linear way or ends up in a kind of stationary state.

The final goal is to achieve the ability to say with certainty that, having trained the machine on *acq.1* at the start of the week, then testing the machine with *acq.(1+x)* will give as a result an accuracy level that decreases with some tendency depending on *x*, where acquisitions taken throughout the day are ordered as *acq1*, *acq2*, etc…

In this work we want to explore this tendency, to see whether
(a) it exists
(b) it is linear or nonlinear.

# IV. Materials and Methods

This section contains detailed descriptions of the dataset, the feature extracting methods and the classifier chosen for the experiments, and a description of their configuration.

**A. Dataset**

Studying repeatability needs a suitable dataset, where measurements of postures have been taken repeatedly several times. According to this, the dataset used contains acquisitions from a four day period.
Acquisitions were taken three times per day with a gap of 2 hours. Approximately the first acquisition was at 10:00 am, then the second within 12:00-13:00 and the last one around 14:00. In this way the influence of disturbing elements that occur in a normal daily routine have been recorded and studied. This leads to a more realistic view of the phenomenon.

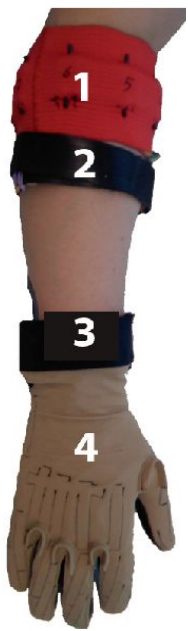

Data have been gathered using several surfaces EMG sensors (see Fig.3), designed to record hand kinematics (i.e. position of the fingers, hand and wrist joints), dynamics (i.e. forces exerted at the fingertips) and the corresponding muscular activity. The sensors were connected to a laptop responsible for data acquisition. Muscular activity has been gathered using ten active double-differential OttoBock MyoBock 13E200 sEMG electrodes, which provide an amplified, bandpass-filtered and rectified version of the raw sEMG signal. These electrodes were fixed on the forearm using an elastic armband. Particular care was taken in the placement of the electrodes on the forearm, since this is usually regarded as a crucial step for data usability. For this reason two methods common in the field were combined: a dense sampling approach and a precise anatomical positioning strategy.

_FIG.3(credits: [2] )Worn sensors: 1. Equally spaced electrodes 2. Spare electrode 3. Inclinometer 4. Cyberglove_

A set of 17 hand and wrist movements of interest was defined including: 8 isometric, isotonic hand configurations and 9 basic movements of the wrist. They can be found in Table I below.

TABLE I

| HAND POSTURES | WRIST MOVEMENTS |
|---|---|
| 1 Thumb up<br>2 Flexion of ring and little finger; thumb flexed over middle and little<br>3 Flexion of ring and little finger<br>4 Thumb opposing base of little finger<br>5 Abduction of the fingers<br>6 Fingers flexed together<br>7 Pointing index<br>8 Fingers closed together | 1-2 Wrist supination and pronation (rotation axis through the middle finger)<br>3-4 Wrist supination and pronation (rotation axis through the little finger)<br>5-6 Wrist flexion and extension<br>7-8 Wrist radial and ulnar deviation<br>9 Wrist extension with closed hand |

Fig.4 below graphically shows each movement:

Hand postures (labels from 1 to 8):   Wrist movements (labels from 9 to 17):

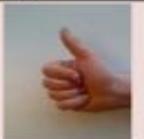
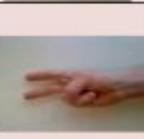
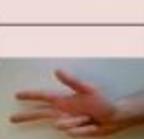
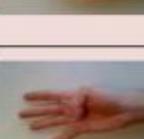
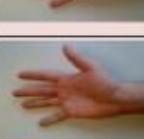
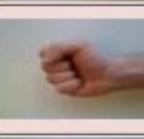
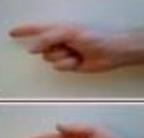
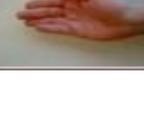
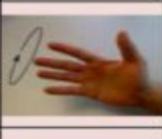
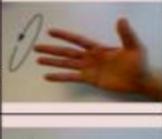
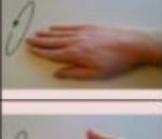
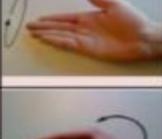
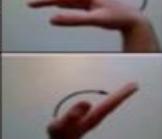
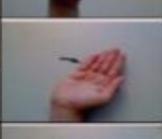
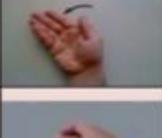
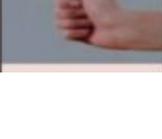
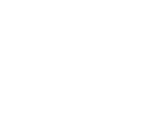

*FIG.4 Set of movements. Pictures taken from [9]*

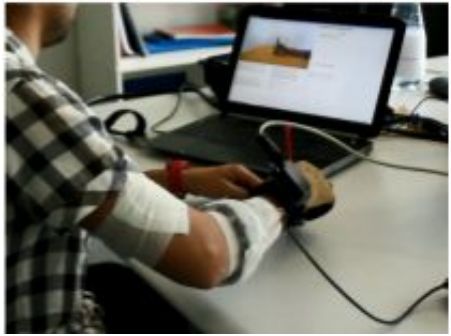

As for the acquisition procedure, the subject is on an adjustable chair in front of a large screen. The electrodes are worn on the right hand. A short movie appearing on the screen presents the movement that should be replicated as accurately as possible. A sequential set of ten repetitions for each class of movements is presented to the subject while data are being recorded. Each movie lasts five seconds; three seconds of rest are allowed in between movements.

*FIG.5 Acquisition setup for movement acquisition. Photo credits: [10]*

### B. Feature Extraction

The choice of methods for features extraction stems from several assumptions on sEMG (see [1] and references therein). For instance, we consider that there is a quasi-linear relation between Root Mean Square amplitude of signal and force exerted by a muscle. As far as this assumption holds, time domain features, such as Mean Absolute Value (MAV) or Waveform Length (WL), when treated in multi-channel settings, could potentially encode profile of movement through force-related measurements.

In addition, we assumed that sEMG can be modeled as a summation of Motor Unit Action Potential trains and that sEMG spectral characteristics might be related to conduction velocity of muscle fibers, which are subject to a great number of conditions [13]. These considerations are related to time-domain features representations.

On the basis of this speculation, two kind of feature representations have been taken into account. All the features are computed from signal $x$ of length $T$ and subindexed by $t$.

### B.1 Waveform Length

Features in the time domain are generally quickly calculated because they do not need a transformation [3].

Waveform length is the cumulative length of the waveform over the time segment.

It is defined as

$$\hat{x} = \frac{1}{T} \sum_{t=1}^{T} (x_t - \bar{x})^2$$

This parameter gives a measure of waveform amplitude, frequency, and duration all in one. To make it clearer, picture below show the raw data representation for the movement 'Thumb up' and the WL related feature.

Raw-signal                                                          Waveform Length

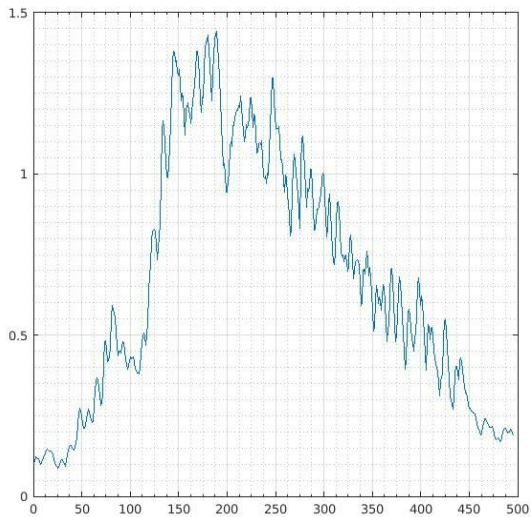
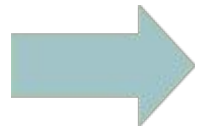
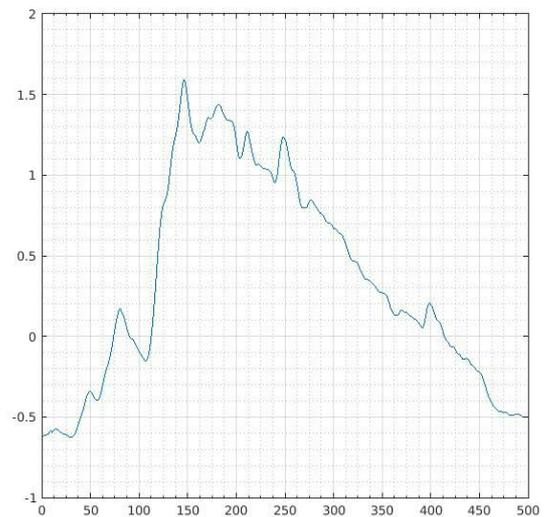

*FIG.6* *Starting from raw data to WL feature representation of the signal. Much of the principal information is mantained.*

### B.2 Short-Time Fourier Transform

Time-frequency representation can localize the energy of the signal both in time and frequency, thus allowing a more accurate description of the physical phenomenon.
The Fourier transform contains high accuracy details in frequency domain, which is optimal for stationary signals. In order to avoid the loss of time related information the STFT multiplies the signal for a sliding window function, $g$. Therefore, STFT has a fixed tiling. Once specified each cell has an identical aspect ratio. We consider $M$ frequency bins indexed with $k$ and computed over blocks obtained by this sliding window function $g$ of length $R$.

The STFT is defined as

$$\hat{x}_{k,t} = \sum_{m=0}^{R-1} x_{m-t} g_m e^{-i\frac{2\pi}{M}km}$$

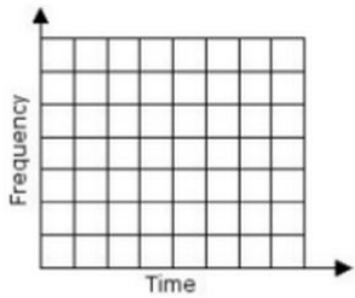

The main constraint of this feature extracting method is that each cell in the time-frequency domain plane must have identical shape. In fact, the plane is divided into cells of temporal width $T$ and frequency height $F$.

Clearly the energy distribution of physical signals is not conveniently localized in regions of fixed aspect ratio.

FIG. 7 Tiling in STFT (figure credits to [3])

Time-Frequency domain features are richer in details than their time domain counterparts, as is shown in pictures 7a and 7b for movement 'Thumb up' with respect to WL representation.

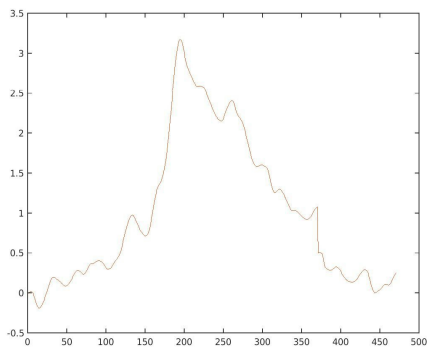

(a)

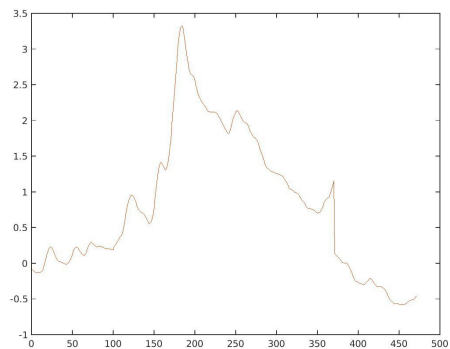

(b)

FIG.8a STFT representation for movement 'Thumb up'

FIG. 8b WL representation for movement 'Thumb up'

## C. Classification

Classification is the process by which we want to assign a label to each sample in the input space. The machine learning method we examined need to be first trained on a set of points in the input space for which the target (label value) is known. This set will be called *training set.* Then, in order to verify that the obtained models are good, they are tested on a separate group of points called *testing set*.

Support Vector Machines (SVM) are learning systems that use a hypothesis space of linear functions in a high dimensional feature space. They are linear binary classifiers that attempt to maximise the distance of two classes to an hyperplane that linearly separates them [6]. Their widespread popularity is due to large extent to the possibility to use kernel functions, which are the key to the efficient use of high dimensional feature spaces. Although they are defined in a binary form, through the conversion of multi-class classification problems into multiple binary problems they allow even multi-class analysis.

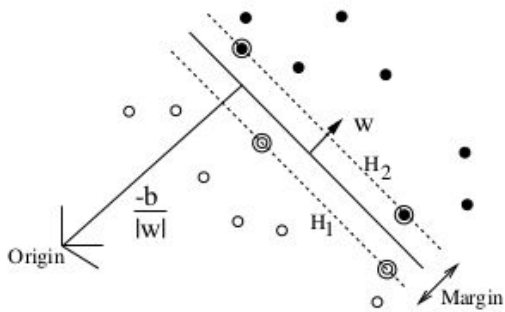

In linear discrimination there is a function that searches for an hyperplane that correctly divides two training data classes (see Fig.9 on the left). In a one-vs-one approach two classes are compared with each other and the best hyperplane that splits data is found through an algorithm from optimisation theory.

<u>FIG.9</u> *Hyperplane that linearly separates data for the separable case. Support vectors are the ones circled. Credits to [11].*

We used SVM with Radial Basis Function (RBF) Gaussian kernel in one-vs-one multiclass classification setting.
Kernel functions are used to define the implicit feature space in which the linear learning machine operates. RBF is one of the basic kernels often used in classification.

This classification method needs the setting of hyperparameters C and $\gamma$, that have been tuned for each experiment by grid search, according to directions in [1].
During grid search we consider non-linear SVM C between $2^i : i \in \{0, 2, ..., 14, 16\}$ and RBF $\gamma$ between $2^i : i \in \{-16, -14, ..., -4, -2\}$
To solve the SVM optimization problem with RBF kernel we have used LibSVM [15].

# V. Experiments

The aim of the experiments is to show how postures classification changes during the week. Between the acquisitions considered there is a 24 hours period and the sensors have been removed during the night. As a result, decrease of accuracy could be due to possible shifts in the electrodes displacement after reapplication. The acquisitions have been ordered and divided into four groups of three, one set for each day. We trained the classifier with the acquisitions of Day 1. Firstly, we trained with the data acquired in the morning and tested on the other morning acquisitions of Day 2, 3 and 4 (see Table II, dataset 1)**.** Subsequently, we repeated the same study on the second dataset containing acquisitions of middle morning (12:00) and finally on the last ones taken around 14:00.

TABLE II

|  | TRAINING | TESTING |
|---|---|---|
| **DATASET 1** | DAY 1 early morning | DAY 1, DAY 2, DAY3 and DAY 4 early morning |
| **DATASET 2** | DAY 1 mid morning | DAY 1, DAY 2, DAY3 and DAY 4 mid morning |
| **DATASET 3** | DAY 1 early afternoon | DAY 1, DAY 2, DAY3 and DAY 4 early afternoon |

All the considerations are related only to a single subject. Considering that all forearms are different with each other in shape, size and power , each conclusion should be taken as a preliminary observation and not as a recurring behaviour.

### A. Experimental Setup

The acquisition consisted of 10 guided 5 seconds repetitions for each of the 17 postures with an allowed period of 3 seconds of rest between movements. The total duration of each acquisition is about 20-30 minutes.

We employed a control scheme already existent in the literature, consisting of preprocessing the signals, segmenting them in windows, subsequently extracting features from the windows, and finally classifying the extracted features. These phases will be detailed in the following subsections.

### A.1 Signal Processing

Several signal processing steps, which are briefly shown in Fig.10, were performed on raw data in order to increase the overall performance. They are described as follows.

*Synchronization*: To synchronize the data streams during acquisition were used high-resolution timestamps. All data are synchronized by linear interpolation to the highest recording frequency (100 Hz).

*Relabelling*: The movements performed by the subject are slightly shifted and not perfectly matching with the stimulus proposed by the video, since some time to react is needed. As a result there is a percentage of label "noise" in the data that has been corrected with an offline relabeling algorithm. For instance, samples with an ambiguous label, that is those recorded during transition between rest and the actual movement, are removed by dividing each movement (including rest) in three equally sized segments and only retaining data from the center segment (see [1] for further details).

*Filtering*: The electrodes are not shielded against power line interferences, thus prior to features extraction the sEMG signals are low pass filtered at 1 Hz using a zero-phase second order Butterworth filter, following the successful configurations adopted in [1, 2, 9]

*Windowing*: After filtering, each signal channel is segmented into windows. Windows of length 100ms and 200ms have been taken into consideration. Between windows there is an overlap of N - 10ms, where N is the length of the window. A longer window could improve the level of detail, while increasing the computational time [1].

*Features Extraction*: Among the methods for features extraction using sEMG signals the ones considered in the experiments are Waveform Length (WL) for time domain analysis and Short-Time Fourier Transform (STFT) as a more sophisticated type. Being STFT in time-frequency domain, the resultant analysis contains richer information at the cost of an increased computational time. Features have been extracted from each window independently for each electrode channel. Based on preliminary evaluation runs we selected a 4-sample rectangular window for STFT. Apart from that, WL did not require other explicit parameters.

*Splitting*: The dataset is split equally into training and testing set at a 50% ratio. In other words, 5 repetitions for each class of movements have been taken as training and the remaining 5 have been included in the testing set. In order to achieve a computationally feasible training set, the training set is subsampled by 10%, meaning that only every 10th sample of a class has been kept. Subsequently, from the testing set it has been taken another 20% for the validation set, that has been used for tuning the hyperparameters of the classifiers. For instance, from the 5 repetitions for each movement in the testing set, the first repetition has been included in the validation set, while the other 4 have been kept into the testing set.

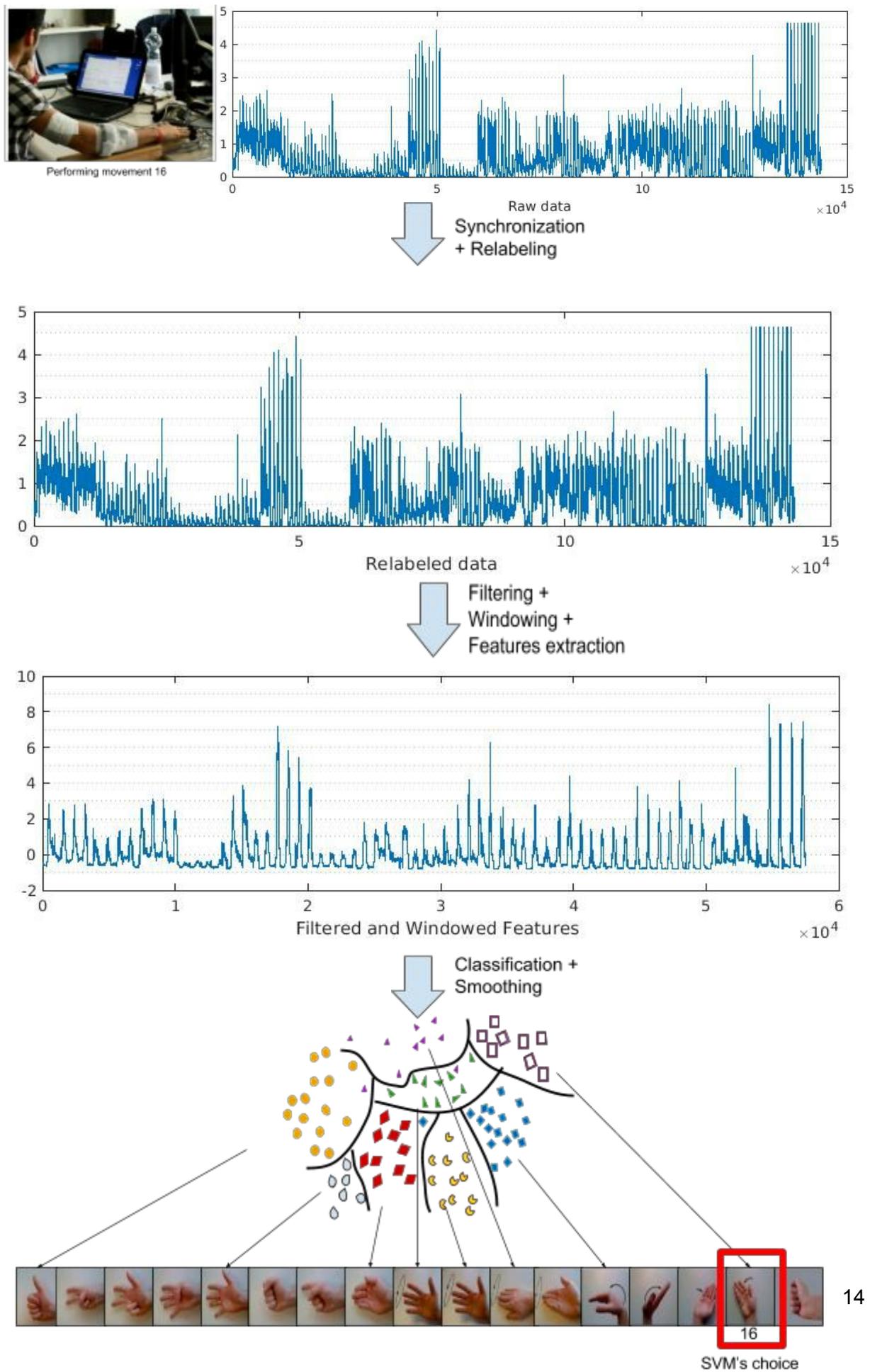

*FIG.10 Experiments data workflow. Credits for acquisition setup photo to [10].*

## A2. Method Configuration and Implementation

We used SVM with RBF kernel in a one-vs-one classification setting.

We tuned separately for each experiment the hyperparameter C and $\gamma$ using grid search. Instead of n-fold cross-validation we splitted the dataset into training/testing/validation sets, speeding up computation time. To build the validation sets we sticked to two methods, that divided the experiments into two different parts. At first, we tuned the hyperparameters using some data from future target acquisitions. In this case the validation set was composed by the sum up of the 20% from each future target acquisition. For example, if we had as training acquisition the first one of Day 1, and we wanted to test the svm on the first acquisitions of Day 2, 3 and 4, then we made our first-part validation set with 20% from test set of the first acquisition from Day 1 plus 20% of the first from Day 2 plus 20% from Day 3 plus 20% from Day 4. In this way we tuned the classifier with some hints of the future target labels to classify expecting higher accuracy results.

Subsequently, we removed all future hints and shrinked the validation sets only to 20% of the training acquisition. To make a parallel in application field, with this two methods we wanted to study what happened training the prosthesis at morning with some hints of the subject future conditions and, with the second one, we wanted to show how an off-line classifier trained only in the morning, reacting to postures throughout the week.

To sum up, at each grid point the SVM has been tuned with a validation set made of:
- I configuration: 20% of the first + 20% of the second + 20% of the third + 20% of the fourth testing acquisition as far as the first part of experiment is concerned (i.e. experiments 1, 2 and 3. See Table III below)
- II configuration: only 20% of the target acquisition in the second part. (i.e. experiments 4, 5, 6)

TABLE III

Configuration method for each experimental dataset.

| CONFIGURATION | I (20% + 20% +..) | II (only 20%) |
|---|---|---|
| EXPERIMENT N. | 1, 2, 3 | 4, 5, 6 |

Although the hyperparameters were tuned for each experiment, in Table IV we present some recurring values.

TABLE IV

Hyperparameters recurring values that emerged throughout the experiments

| hyperparameter | WL | STFT |
|---|---|---|
| $C$ | -2 | 4 |
| $\gamma$ | -4 | -8 |

Further details on hyperparameters and configuration methods are widely reported in the Appendix sections A1 and A2.

**A3. Post classification smoothing**

Final accuracy can be significatively enhanced with a little additional cost by computing a simple smoothing algorithm. We applied this algorithm only one time but smoothing can be done several times, until a significant increment in accuracy can be found. In this way it is possible to highlight only the dominant, thus presumably correct, classification.

Thanks to smoothing we managed to reduce misclassification of movements reaching an edge of 90.5% of accuracy with a maximum gain of a 4%.
The aim of this technique is to give a general idea of relatively slow changes in the predicted labels, which are numbers that identify the class of appartenance of the movements recognized from the classifier. We implemented a sliding window of a variable length that finds the most frequently predicted label in a majority vote approach. Once found the most popular value we approximated all the values contained into the window to that one. As for window length parameter, we have chosen the one with the maximum increment of accuracy.
An example of smoothing applied to classification predicted labels can be found in picture 9. It is clearly visible that it eliminates a great part of misclassification errors.

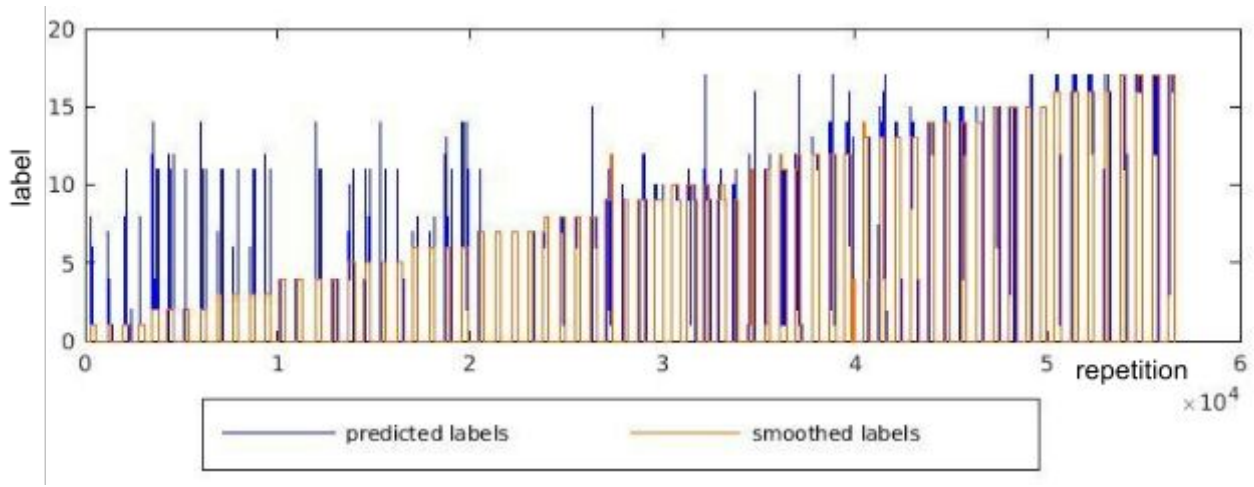

*FIG.11 Smoothing of the predicted labels for experiment 1. Yellow line is the misclassification error correction, and it is clearly visible that all the blue spikes corresponding to the mistaken predicted labels are successfully corrected by smoothing technique.*

Every step has been implemented through MATLAB interface with the support of Statistics Toolbox and LibSVM [15].

## B. Results

Here are presented evaluation results in three perspectives: classification accuracies according to features, single movements classification accuracies and misclassification analysis by confusion matrices.

### B1. Classification accuracy tendency

Classification accuracies for the methods considered in section IV with respect to feature representations Waveform Length and Short-Time Fourier Transform are summarized in an overall graph, showing both the non smoothed and the smoothed results.

In all the following graphs we will refer to Waveform Length as WL specifying if it is the not smoothed result (WL-nS) or the smoothed one (WL-S). Same considerations have to be applied to Short-Time Fourier Transform (STFT-nS and STFT-S).

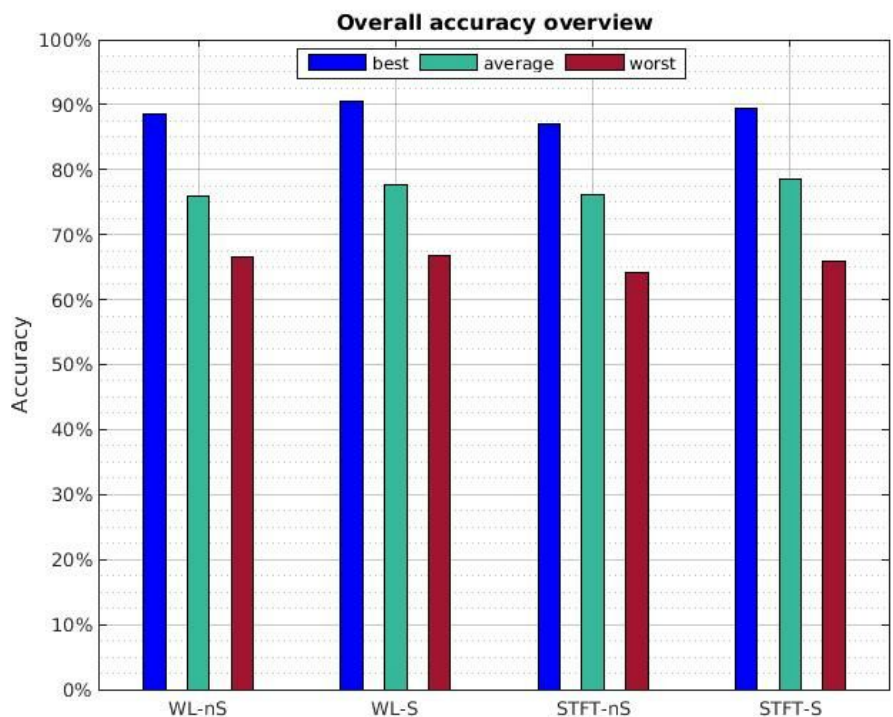

*FIG.12 Classification accuracies. Each bar represents method classification accuracy with respect to feature representation and smoothing.*

The best performing results have been obtained with WL after smoothing, reaching accuracies higher than 90%. Best results are related to the testing acquisitions that immediately followed the training sessions. This means that at the beginning of the week, after the training session, the number of successfully recognized gestures among the 17 considered is between 15-16. The worst performance at all, as we had predicted, is the one with the acquisitions taken at the end of the week. In this case, even choosing the most performant combination of methods and techniques, the amount of postures decreases to 10-11. Most of the time the algorithm offers a number of recognized gestures between 12-13. This shows, and it is even more visible in the following pictures, that the deterioration of accuracy highly impacts on the results, but even in the worst case scenario it will be responsible for a maximum loss of 5 movements, which is less than a 30% of the whole set of movements.

What we want to highlight now is a general tendency in performance to stabilize around 70% of accuracy progressively throughout the week.

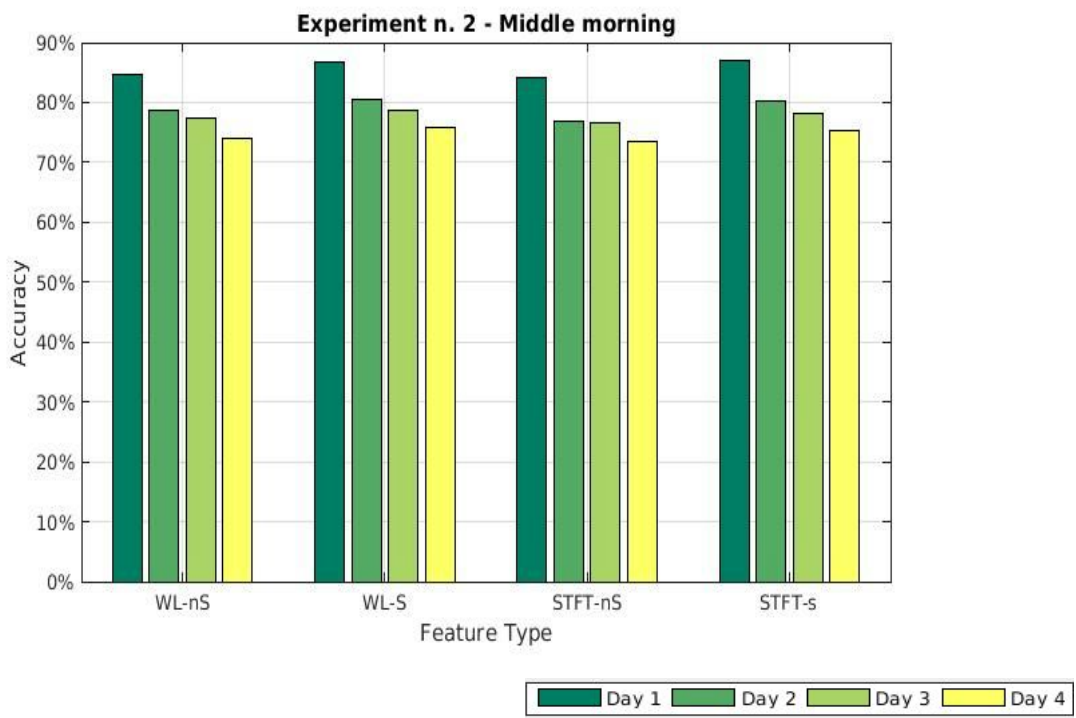

*FIG.13 Testing accuracies with training dataset from day 1 at 9:00 a.m. and SVM tuned with the first configuration of the validation set.*

These pictures (13, 14, 15) are some of the most significant and recurrent tendencies that emerged during the experiments.

The accuracies are showed in groups divided by feature-type and each bar is the result of testing sessions taken during subsequent days.

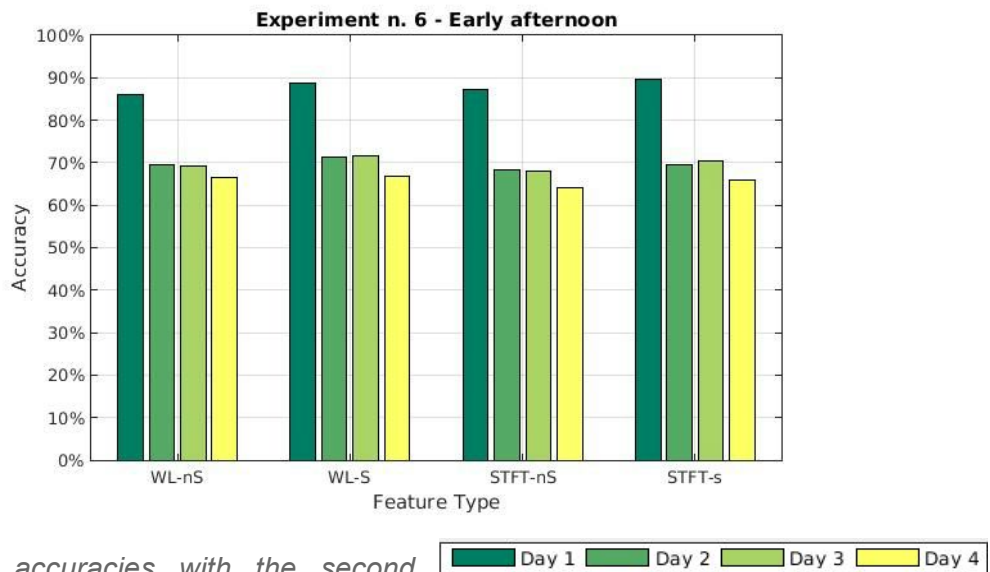

FIG.14 *Testing accuracies with the second configuration of the SVM tuning set trained on 2:00 p.m. acquisitions.*

We see that the higher bar is always the one related to the testing set taken from the same acquisition of the training set. The bars that follow in different shades of green are the accuracy of testing the algorithm during the following days.

However, all the obtained results cannot be intended as a fixed rule. In fact, Fig. 15 shows an unpredicted rise of accuracy that occurred in experiment n.4 (which is relative to training the SVM only on the first morning acquisition). Even in this case it is important to remark that the accuracy drop tends to stop around 70%.

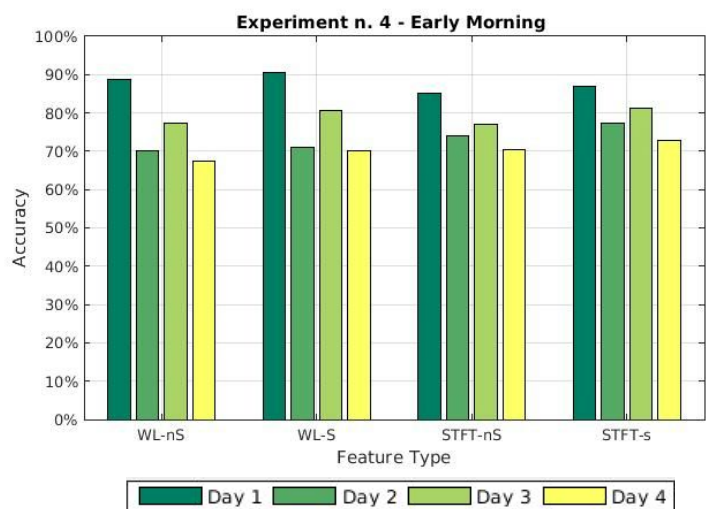

FIG.15 *Testing accuracies with the second configuration of the SVM tuning set trained on 9:00 a.m. acquisitions.*

## B2. Single movements classification accuracies

We now present a detailed analysis of the most frequently recognized movements.

It emerges that wrist extension has reported the highest accuracy level, while the abduction of all fingers is the movement correctly classified most of the times with average levels of accuracy. The worst accuracy emerged overall during classification of the hand posture 'Thumb up'.

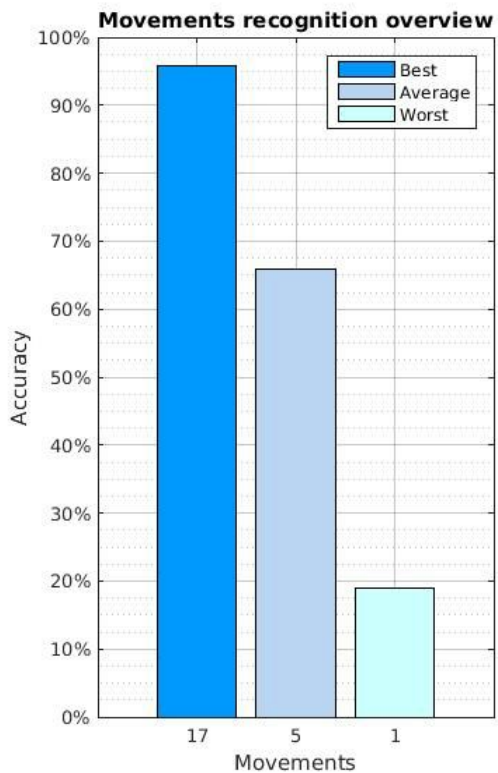

Movement 17:

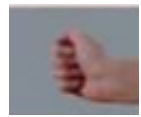

closed hand with wrist extension

Movement 5:

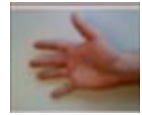

all fingers abduction

Movement 1:

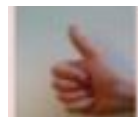

thumb up

*FIG.16 Single movements analysis. On the left histogram are presented the highest reported accuracy (in deep blue) and the worst overall (light blue). On the x axis can be found the movement label, whom corresponding pictures are shown in detail on the right side. Credits for movements pictures to [9]*

The two following pictures show that the classification accuracy for each single posture lays between 80-90% in the most frequently recognized ones. In all the experiments the accuracies related to labels 3 and 14 stand out among the best accuracies. The quick change of accuracy related to posture 7 shows that lack of stability in the correctly predicted labels is still a problem. However, the presence of persistent movements in the higher accuracies as for 3 and 14 let us sense a possible distinction between stable movements that are most of the times surely classified correctly, and movements with an uncertain behaviour.

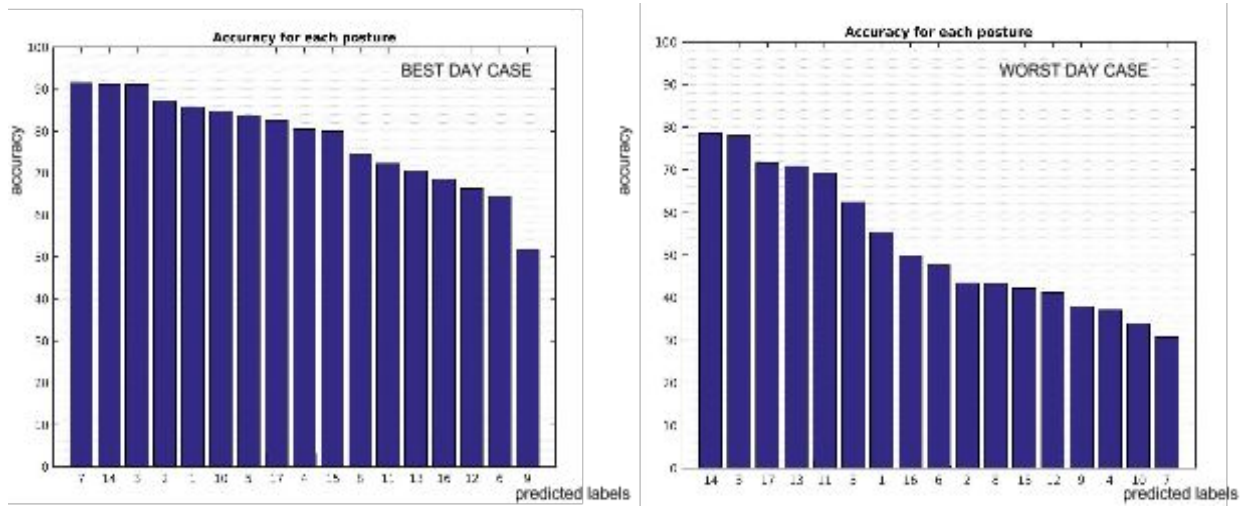

*FIG.17 Singular accuracies related to movements classification taken at different stages of the experiments, i.e. during Day 1 early morning for best case and during Day 1 early afternoon for the worst.*

C. Discussion

From a statistical point of view, if considering all movements and postures equiprobable, the probability of randomly choose the correct label among all the 17 possible hand configurations and gestures is given by

$$\frac{1}{17} \simeq 0.06$$

which is 6%. Taking this into account, the obtained results are surprisingly high enough to guarantee a correct classification of more than 10 postures in each moment of the week. This means that even in the worst case scenario, there is a percentage of success that remains still around 58-60%. For this reason, this study clearly shows that research is going in the right direction, while a lot still has to be done.

# VI. Conclusions

In this work we explored the repeatability task of sEMG signals for the control of advanced non-invasive prosthetic hands. We examined the changes in classification accuracies over a 4 days period. Prior to feature extraction we introduced a preprocessed version of the dataset consisting of several steps such as synchronizing the raw data, relabelling them to correct rest and movement transition time points. Before features extraction, we low pass filtered the signal and segmented it into windows. We then used WL and STFT to represent the datasets.

After doing that, we trained the SVM classifier with the first day measurements dataset adopting two different strategies for tuning the hyperparameters. In the first, we included some hints of the future target movements labels, in the second we used only one acquisition dataset.

We tested the SVM with the signals from each of the three considered times of the day, i.e. early morning (9:00), mid morning (12:00) and early afternoon (14:00), for each of the 4 days.

Classification results, having a 60% lower boundary for accuracy, revealed that the steps done in this field are in the right direction. We observed that there is a tendency to stabilize around the 10 successfully classified movements through the days, starting from an encouraging number of 14-15 recognized postures right after the training session. It did not emerged a fixed rule in the decrease of accuracy. As shown in experiment n.4, results can be slightly variable.

Moreover, we saw that there are two groups of movements that behave in different ways. While one of them has a stable line, that can be really high or really low in each of the 6 experiments, the other one is much more unstable, turning from high degrees of accuracy to really bad performances.

Taken all the previous considerations into account, we conclude that performances heavily decrease after the first day but remain in the same range the days after, letting the intuition of a stability level. In other words, this means that there is no relevant difference in performance after 1, 2 or 3 days.

## A. Future work

It is important to say that the obtained results are still far away from what can be considerably usable in real-life settings. We wanted also to underline that this is a pioneer pilot study with only one subject acquisition dataset of four days. Therefore, it is not possible to consider these results as fixed rule because the number of experiments is still not statistically significant.

For this reason, possible directions for future work can be testing the observed behaviour on an extended dataset. Dataset can be widely extended in many possible directions, which can be for instance increasing the number of subjects or the long term period from 4 days to one week or even more.

Another possible research thread could be to study how performance might increase and/or get stable, when using multiple feature types, combined together with principled classifiers [16]. Lastly, the probable need for a periodic recalibration of the device might be made less intense for users by combining it with adaptive algorithms able to significantly shorten the training time [17].
Future work will focus on these directions.

# IX. Appendix

In the appendix we present in the first part (A) the experimental setup details, with particular attention to the acquisition schedule and to the database settings. In the second part (B) we present further analysis of the results, including the best and worst case overview and misclassification matrices. In section B3 we report the accuracy values of the experiments. Following we will refer to acquisitions with the term acq. and to experiments with exp..

## A. Experimental setup details

In this section we give detailed information on the acquisition scheduling and dataset organization for each of the experiments.

### A1. Acquisition schedule

We report the acquisition schedule in Table V to clear up how the acquisition were distributed throughout the days. We used 12 acquisition files. We numbered them starting from acq.2 to acq.14. Please note that we skipped acq.1 and acq.4 because the files were corrupted.

TABLE V
Acquisition Schedule

|        | DAY 1    | DAY 2    | DAY 3     | DAY 4     |
|--------|----------|----------|-----------|-----------|
| **9:00**  | acq n.2  | acq n.6  | acq n.9   | acq n.12  |
| **12:00** | acq n.3  | acq n.7  | acq n.10  | acq n.13  |
| **14:00** | acq n.5  | acq n.8  | acq n.11  | acq n.14  |

Given this, in the following table we show the acq. used to build in the experimental datasets.

TABLE VI
Experimental datasets basic structure.

| EX number     | 1 and 4 | 2 and 5 | 3 and 6 |
|---------------|---------|---------|---------|
| *Training acq.* | 2       | 3       | 5       |
| Testing acq.  | 2       | 3       | 5       |
| Testing acq.  | 6       | 7       | 8       |
| Testing acq.  | 9       | 10      | 11      |
| Testing acq.  | 12      | 13      | 14      |

## A2. Datasets setting

Table V and Table VI show that exp.4, exp.5 and exp.6 used datasets that are similar respectively to the ones used in exp.1, exp.2, exp.3. Thought the couples of experiments 1 and 4, 2 and 5, 3 and 6 used the same measurements, the difference between them lays in the setting of the splitting function. For this reason, we divided the experiments into 2 parts. Part 1 contains exp.1,2,3 while Part 2 exp.4,5,6. Experiment splitting is summarized in Table VII. As we said in V.A2, we tuned the hyperparameters using one method in Part 1, that consisted in tuning the classifier with some hints of the future conditions, while in Part 2 we removed them all.

TABLE VII
Experiments in Part 1 and Part 2

| Part 1 (Configuration I) | Part 2 (Configuration II) |
|---|---|
| exp. 1, 2, 3 | exp. 4, 5, 6 |

In brief, with configuration I we refer to datasets with the validation set build with the sum up of the 20% from each of the target acquisition testing set, while configuration II is for the ones with only 20 % from the training acquisition test set. Details are presented in Table VIII.

TABLE VIII
Configuration used for each experiment.

| EXPERIMENT N. | 1 | 2 | 3 | 4 | 5 | 6 |
|---|---|---|---|---|---|---|
| DATASET N. | 1 | 2 | 3 | 1 | 2 | 3 |
| CONFIGURATION | I | I | I | II | II | II |

## B. Further results analysis

In this section, we present, more in detail, the accuracies reported during the experiments. Firstly, we make a comparison between the best case scenario and worst one. Secondly, we analyze misclassification with confusion matrices and finally we present all the accuracy values.

### B1. Best/Worst case comparison

In the following we want to compare the best case scenario with worst case scenario accuracies. Bar graphs show that there is an interval of approximately 20% between the best blue bar and the best red bar. This interval is kept both in the average and in the worst case of the two scenarios.

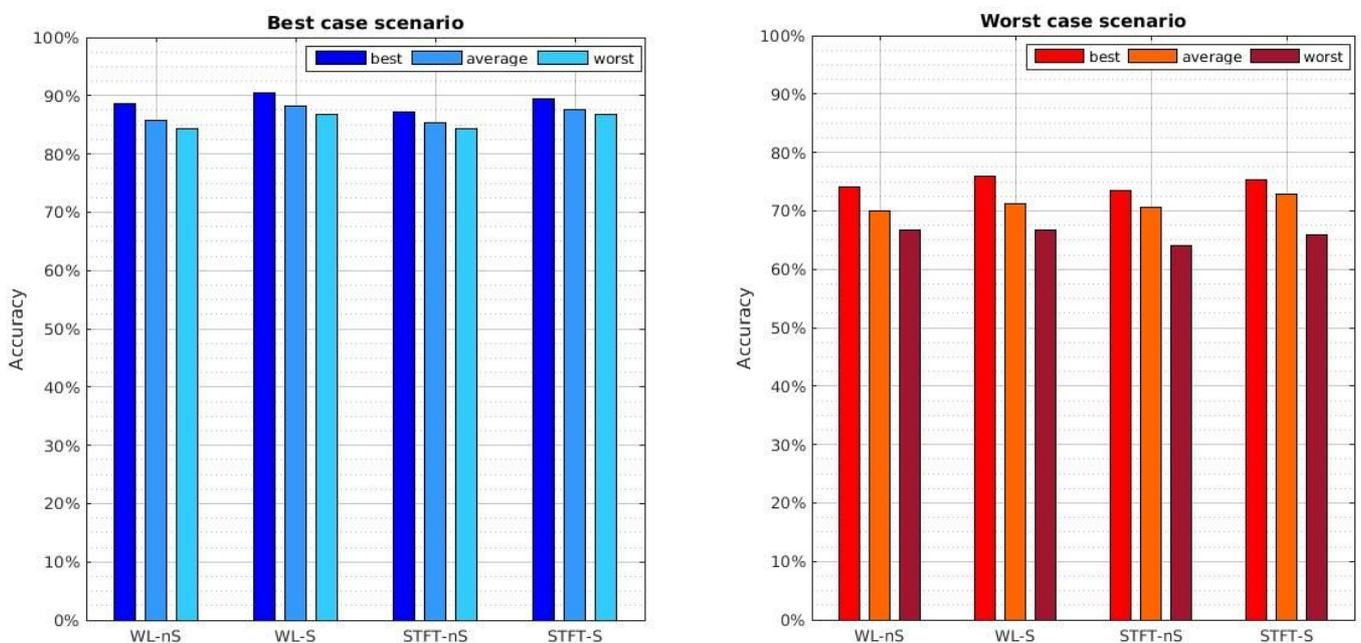

*FIG. 12 In best case scenario accuracy rises over 90% levels in the most performant combination of techniques. It is important to see that even in the worst case accuracy never decreases under 60%. In other words this means that at least 10 successfully classified movements are guaranteed over the week.*

## B2. Misclassification

The confusion matrices taken over the week show principally that misclassification increases as more movements start to be confused with others.

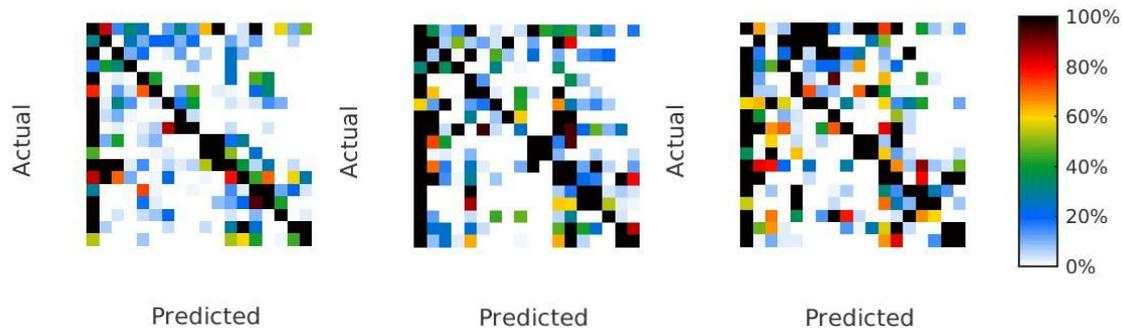

*FIG.18 Confusion matrices for the SVM with WL feature representation without smoothing. Each cell represents prediction accuracy of row indexed class. First class is rest, i.e. absence of movement. Correct predictions would result in clear left-top to right-bottom diagonal. Off diagonal cells are indicative of misclassification.*

## B3. Accuracy values

Table containing accuracy results can be found below.

|        | Train | Validation | Test | Accuracy - WL | | Accuracy - STFT | |
|--------|-------|------------|------|---------------|---|-----------------|---|
|        |       |            |      | non Smoothed | Smoothed | non Smoothed | Smoothed |
| Part 1 | 2 | 2,6,9,12 | 2 | 84,41 | 86,76 | 84,88 | 86,72 |
|        |   |          | 6 | 73,22 | 74,91 | 73,97 | 76,39 |
|        |   |          | 9 | 77,4  | 78,59 | 77,5  | 79,5  |
|        |   |          | 12 | 71,4 | 72,82 | 71,91 | 74,51 |
|        | 3 | 3,7,10,13 | 3 | 84,57 | 86,8 | 84,25 | 86,91 |
|        |   |          | 7 | 78,62 | 80,55 | 76,94 | 80,29 |
|        |   |          | 10 | 77,32 | 78,73 | 76,59 | 78,24 |
|        |   |          | 13 | 74,07 | 75,85 | 73,5 | 75,33 |
|        | 5 | 5,8,11,14 | 5 | 85,63 | 88,83 | 86,5 | 88,74 |
|        |   |          | 8 | 70,27 | 70,85 | 73,9 | 76,52 |
|        |   |          | 11 | 70,25 | 71,97 | 73,6 | 74,8 |
|        |   |          | 14 | 67,71 | 67,08 | 69,9 | 73,55 |
| Part 2 | 2 | 2 | 2 | 88,61 | 90,49 | 85,17 | 87,03 |
|        |   |   | 6 | 70,22 | 70,99 | 73,98 | 77,26 |
|        |   |   | 9 | 77,21 | 80,73 | 77,12 | 81,17 |
|        |   |   | 12 | 65,51 | 69,98 | 70,4 | 72,74 |
|        | 3 | 3 | 3 | 85,49 | 87,45 | 84,25 | 86,91 |
|        |   |   | 7 | 77,13 | 79,06 | 76,94 | 80,29 |
|        |   |   | 10 | 77,03 | 79,88 | 76,59 | 78,24 |
|        |   |   | 13 | 72,88 | 74,58 | 73,5 | 75,33 |
|        | 5 | 5 | 5 | 86,09 | 88,72 | 87,13 | 89,52 |
|        |   |   | 8 | 69,46 | 71,4 | 68,17 | 69,4 |
|        |   |   | 11 | 69,17 | 71,68 | 68,06 | 70,28 |
|        |   |   | 14 | 66,62 | 66,76 | 64,06 | 65,85 |